\documentclass[letterpaper, 10 pt, journal, twoside]{IEEEtran}
\usepackage{amsmath,amsfonts}
\usepackage{algorithm}
\usepackage{algpseudocode}
\usepackage{array}
\usepackage[caption=false,font=normalsize,labelfont=sf,textfont=sf]{subfig}
\usepackage{textcomp}
\usepackage{stfloats}
\usepackage{url}
\usepackage{verbatim}
\usepackage{graphicx}
\usepackage[sorting=none,doi=false,url=false]{biblatex}
\usepackage{graphicx}
\graphicspath{ {./images/} }
\usepackage[acronym]{glossaries}
\usepackage{optidef}
\usepackage{hyperref}
\usepackage[capitalise]{cleveref}
\usepackage[figurename=Fig.]{caption}
\usepackage{subfig}
\usepackage[table]{xcolor}
\usepackage{tkz-euclide}
\usepackage{pgfplots}
\usepackage{authblk}
\pgfplotsset{width=7cm,compat=1.9}
\bibliography{mybibliography}

\title{Addressing Ambiguity in Imitation Learning through Product of Experts based Negative Feedback}

\author[1]{John Bateman}
\author[1]{Andy M. Tyrrell}
\author[1]{Jihong Zhu}
\affil[1]{University of York}

\begin{document}

\maketitle

\begin{abstract}

Programming robots to perform complex tasks is often difficult and time consuming, requiring expert knowledge and skills in robot software and sometimes hardware. Imitation learning is a method for training robots to perform tasks by leveraging human expertise through demonstrations. Typically, the assumption is that those demonstrations are performed by a single, highly competent expert. However, in many real-world applications that use user demonstrations for tasks or incorporate both user data and pretrained data, such as home robotics including assistive robots, this is unlikely to be the case. This paper presents research towards a system which can leverage suboptimal demonstrations to solve ambiguous tasks; and particularly learn from its own failures. This is a negative-feedback system which achieves significant improvement over purely positive imitation learning for ambiguous tasks, achieving a 90\% improvement in success rate against a system that does not utilise negative feedback, compared to a 50\% improvement in success rate when utilised on a real robot, as well as demonstrating higher efficacy, memory efficiency and time efficiency than a comparable negative feedback scheme. The novel scheme presented in this paper is validated through simulated and real-robot experiments.
\end{abstract}
\section{Introduction} \label{introduction}

Imitation learning is an attempt to mimic the form of learning used naturally by humans and many animals \cite{biologicalImitation} in a machine learning context, imitating demonstrations of a task in order to learn to perform it \cite{imitationsurvey}. With the general trend towards more common usage of machine learning, and the applicability of imitation learning to a number of tasks, such as robotics for assisted dressing \cite{assistedDressingTRO} and assisted feeding \cite{liu2024adaptivevisualimitationlearning}, cooperative assembly tasks \cite{sasagawa2020cooperation} and robot locomotion \cite{peng2020learning}, close interaction between humans and robots is becoming more common. In these cases, allowing a human to have control over the robot is an important step to their implementation as the human user should be able to demonstrate the task which they want the robot to perform.

However, whilst there are a number of different methods for acquiring the data, such as teleoperation or motion capture, they are not necessarily intuitive for an inexperienced demonstrator, such as someone demonstrating new behaviour for a robotic system they are utilising. Most imitation learning systems are designed on the assumption that demonstrated data will be relatively optimal \cite{imitationsurvey}, but given the challenges associated with some of the methods of demonstration \cite{fang2019survey}, particularly in the field of assistive robotics, as discussed below, this assumption does not necessarily hold.

For imitation learning to be effective, many systems require the user to be physically capable of performing the task in order to demonstrate it, as many utilise kinesthetic teaching \cite{imitationsurvey}. Since users of assistive technology would require the assistive technology in the first place, a general assumption can be made that, in many cases, they are not physically capable of reliably performing the task optimally. While this can be countered by utilising other teaching methods such as teleoperation, this may not be the most practical way to approach user control over the system. Teleoperation is known to have a significant degrading effect on task performance \cite{kim1992lag}, it often requires a large number of executions to generate an effective alternate policy \cite{imitationsurvey}, and additionally, assistive or collaborative tasks are significantly affected by the human collaborator \cite{sasagawa2020cooperation}. Also, a user new to teleoperating robots or to training this specific task is likely to make mistakes, creating a large set of runs which are discarded rather than being utilised for training. Any failures to perform the task would also be discarded, only giving the user the information that they require more training data. In addition, it has been observed \cite{Camacho1995BehavioralCA} that imitation learning performs at its best with only a single demonstrator, which limits the ability to pretrain a system if user training is required.

In order to enable this kind of training by non-expert demonstrators, a method to robustly incorporate data which is likely to include suboptimal or failed demonstrations is required, and, ideally, one which improves on the policies learned from these demonstrations, in order to achieve a high data efficiency, with improved success rates for a small number of demonstrations. This paper makes a number of novel contributions to the literature, specifically a framework which utilises negative learning in a feedback loop in order to significantly improve success rate of tasks, focusing specifically on ambiguous tasks with very low success rate; a system of negative learning utilising a product-of-experts method to improve the efficiency and effectiveness of this negative feedback loop; and a system to select regions of the trajectory which are unsuccessful based on consensus from successful trajectories.

The rest of the paper is organised as follows: Section \ref{relatedworks} introduces background and related work in resolving ambiguity and negative learning; Section \ref{methodology} presents the methodology for this negative expert feedback algorithm and briefly covers the general structure aimed towards; Section \ref{results} presents results for the experiments with both simulation and real robots; and Section \ref{conclusion} contains the conclusions, a discussion of limitations and future work.

\section{Related Works} \label{relatedworks}
This section covers the necessary background for this paper, as well as covering the related works, in three primary areas - Ambiguity in subsection \ref{ambiguity}, Multiple Experts in subsection \ref{multipleexperts} and Negative learning in subsection \ref{negativeexperts}.

\subsection{Ambiguity} \label{ambiguity} 

Ambiguity is the property of a system which makes it open to interpretation, to differing degrees. In imitation learning specifically, it refers to a group of considerations that introduce challenges in generalising from demonstrations to an actual policy.

As defined in Bensch et al. 2010 \cite{Bensch10Ambiguity}, an ambiguity in imitation learning occurs when the hypothesis space, $H$, contains multiple non-empty hypotheses, $h_i$. Each hypothesis is a possible generalisation of the behaviour of the demonstration. The ambiguity is given as $A = |H|$, that is, $A$ is the size of the hypothesis space.

In this paper, the primary concern is task-based ambiguities - that is, ambiguities arising from the design of the task rather than other sources. This form of ambiguity is illustrated in Figure \ref{fig:ambigSim}, where there is more than one approach to the same task, and the ambiguity of this form of task results in a failure. Multiple expert methods, described in the next section, are a potential avenue to tackling this kind of problem.

Resolving ambiguities in imitation learning has been addressed through several methods. Probabilistic model approaches \cite{ewerton2015mixtureofinteractions} can estimate the uncertainty of a policy in order to find ambiguous situations. When policies have particularly low certainty, they can pass control to a user demonstration \cite{kelly2019hgdaggerinteractiveimitationlearning}, to interactive feedback \cite{Chernova_2009} or to a control algorithm \cite{lee2019safe}. However, while actually performing a task, requesting demonstrations is not necessarily ideal, and while this becomes less likely as the learned policy improves, it could still occur during normal operation if the imitation learning system was being used in, for example, an assistive context. While control algorithms might seem to be superior in this respect, they are also limited in their capacity to actually complete a task, compared to imitation learning, so are primarily used to return the system to a known state \cite{lee2019safe}, which is less helpful in resolving ambiguities as it is in keeping the imitation learning agent away from unexplored regions of space.

Learning Interactively to Resolve Ambiguity (LIRA), proposed in Franzese et al. 2020 \cite{franzese2020learning}, utilises human interaction to resolve task ambiguities and to give the robot awareness of the ambiguities for, specifically, the selection of motion primitive reference frames. However, this also requires direct human intervention to interact with and correct the robot.

\subsection{Multiple Experts} \label{multipleexperts}
Humans generally learn at their best from multiple sources, because we have different learning styles \cite{learningStyles}. This is not true for imitation learning systems, where policies trained on data from multiple experts are typically less effective than those trained by a single expert \cite{Camacho1995BehavioralCA}, due to the ambiguity associated with more complex tasks. However, in many cases, it can be simpler to acquire demonstrations from a larger number of people - and, in the scenario outlined above, the user of the robotic system is likely a different expert than whoever provided the initial demonstrations, necessitating an expert combination method. Several of these exist - Mixture of Experts, proposed by Jacobs et al. \cite{Jacobs1991} creates a policy using a weighted sum of expert policies. Alternately, Product of Experts, proposed by Hinton \cite{Hinton2002} creates the policy by multiplying the policies together and then normalising the result. Product of Experts is popular in reinforcement learning, but has not yet seen much use in Imitation Learning \cite{imitationsurvey}, while Mixture of Experts is more common. Either can be used to resolve ambiguities in tasks \cite{imitationsurvey}, however this requires labeled data, in order to know which demonstration corresponds with which method of task execution, which the proposed method does not.

\begin{figure} [htbp!]
    \centering
    \includegraphics[width=0.9\linewidth]{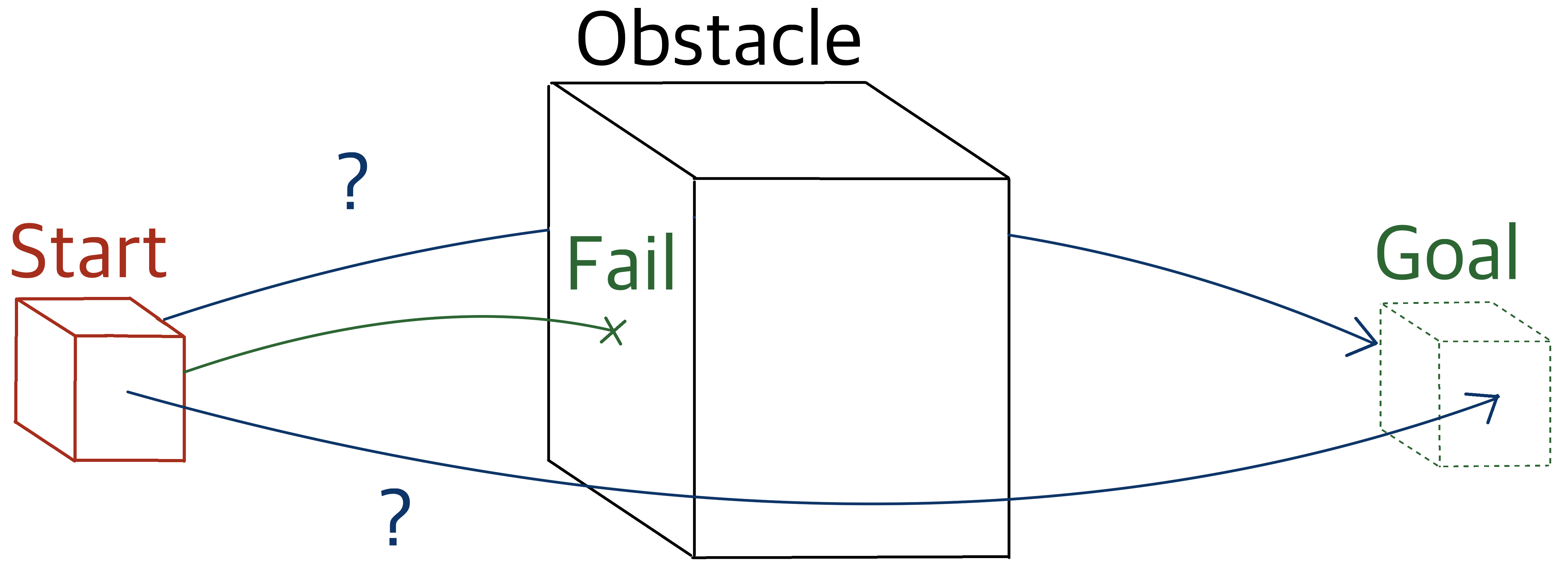}
    \caption{\emph{An illustration of the ambiguity problem in a task avoiding an obstacle while navigating to a goal. The two blue lines represent demonstrations, both successes (neither collide with the obstacle), but the central green line, learned from a combination of the two, is a failure, as it averages the two behaviours and collides with the obstacle.}}
    \label{fig:ambigSim}
    \vspace{-0.5cm}
\end{figure}

Figure \ref{fig:ambigSim} shows the ambiguity problem present in certain tasks - a simple approach to learning this task, without some method of dividing the multiple different expert behaviours, will result in the failed run shown here, a collision between the arm and the obstacle \cite{imitationsurvey}. This failed run is then discarded. 

\subsection{Negative Experts} \label{negativeexperts}
Utilising negative demonstrations - either failures or explicit demonstrations of behaviour to avoid - is a relatively recent innovation in imitation learning, used by Kalinowska et al. \cite{kalinowska2021posneg} and others, although the quality of expert data has been in question in other fields such as supervised learning, where Raykar et al. \cite{raykar2009supervised} utilised negative expert techniques to learn policies from suboptimal data. While typically less effective or outright failed approaches to a task are discarded, there is information to be extracted from them, as it is possible to learn a task purely through negative demonstrations \cite{kalinowska2021posneg}.

One potential benefit of this is learning from mistakes. That is, when a human makes a mistake while performing a task, they learn and improve and avoiding performing the same mistake again. However, imitation learning algorithms do not generally do this \cite{imitationsurvey} - feedback is typically by positive-reinforcement. However, if the algorithm generates runs which are infeasible, whether that is determined by a human observer or automatically by comparison to the task, that data could theoretically be used to avoid making the same mistake twice.

Learning these negative policies individually and then combining them with expert combination methods should also improve the time and memory efficiency of negative learning, particularly negative feedback, since at each instance of negative feedback the system only needs to learn from a single trajectory, and only needs to store the policy representation learned from its training data rather than the training data itself.

\section{Methodology} \label{methodology}

The aim of this work is to present a step towards an overall system which is intended to be capable of robustly learning from failure and from suboptimal as well as correct demonstrations.

The primary contribution of this work is a negative learning algorithm based on product of experts methods - since product of expets combines policies by product, only one of the policies needs to have a low weight on an area of the distribution in order to guarantee a low weight in that same area on the overall distribution. This means that an otherwise uniform distribution with reduced weights on all negative demonstration areas should theoretically be able to minimise the distribution only in regions where negative reinforcement is desired, with limited effects outside of it. If the policy of the final system is a probability distribution \(\pi^*\), there are \(n\) learned policies \(\pi_1\) to \(\pi_n\), and policies learned from negative demonstrations, using the same method as learning from positive demonstrations, and the avoidance policy \(\pi_\alpha\)(which includes policies learned from negative feedback), the proposed system can be represented as

\[ \pi^* = \frac{(U-\mu.\pi_\alpha).\sum_{m=1}^n \pi_m}{\sum_c(U-\mu.\pi_\alpha).\sum_{m=1}^n \pi_m} \]

Where \(U\) is a uniform distribution across the policy space, and \(c\) indexes across that space. The policy combination follows the general form of product of experts combination \cite{Hinton2002}. Ideally, if there are no negative demonstrations then the \(U-\pi_\alpha\) factor becomes the uniform distribution \(U\) and therefore has no impact on expert probabilities. If the avoidance policy can be learned efficiently, with the majority of its probability density around the regions of undesired behaviour, then it should bring the \(U-\pi_\alpha\) probability close to zero in those areas, minimising the probability of those behaviours in the overall policy, with minimal extraneous influence on the positive elements of the expert policies. Learning these policies individually and then combining them with expert combination methods should also improve the time and memory efficiency of negative learning, particularly negative feedback, since at each instance of negative feedback the system only needs to learn from a single trajectory, and only needs to store the policy representation learned from its training data rather than the training data itself.

In this work, we propose a more complete form of this equation,

\[ \pi^* = \frac{\prod_{i=0}^s(U-\mu.\pi_{\alpha i}).\sum_{m=1}^n \pi_m}{\sum_c\prod_{i=0}^s(U-\mu.\pi_{\alpha i}).\sum_{m=1}^n \pi_m} \]

which incorporates $s$ avoidance distributions, $\pi_{\alpha i}$, in a way which allows them to be applied to the final result sequentially and commutatively, allowing multiple iterations of the negative learning to be applied without requiring the original policy or failures to be saved, minimising memory inefficiencies. This method also requires learning from only one failure for each policy $\pi_{\alpha i}$, maximising time efficiency.

Algorithm \ref{alg:feedback} shows the novel negative feedback algorithm proposed in this work; for a dataset $\underline{D}$ of demonstrations, a set of learning policies $\underline{\pi}$, and a trajectory $t_i$ generated from that run of the learning algorithm being used. In this work, the learning method was a Gaussian Mixture Model (GMM) method sourced from Calinon \cite{Calinon07}, using Expectation Maximisation. The Probability Density Function (PDF) was then derived from this GMM and the trajectories were sampled from this PDF.

\begin{algorithm}
\caption{The negative feedback algorithm}
\label{alg:feedback}
\begin{algorithmic}[1]
\State Collect a set of positive demonstrations, $\underline{D}$
\State Learn a policy or set of policies, $\underline{\pi}$, from $\underline{D}$
\State Generate the first trajectory, $t_i$ from $\underline{\pi}$
\While{$t_i$ is not Successful}
    \State Select the desired region of $t_i$
    \State Consider $t_i$ as a negative demonstration dataset $\underline{D}_{Ni}$
    \State Learn avoidance policy $\underline{\pi}_\alpha$ from $\underline{D}_{Ni}$
    \State $\underline{\pi} = (U-\mu.\underline{\pi}_\alpha).\underline{\pi}$
    \State Normalise $\underline{\pi}$
    \State Generate next trajectory $t_{i+1}$ from $\underline{\pi}$
\EndWhile
\end{algorithmic}
\end{algorithm}

The theoretical primary benefit of this algorithm is the efficient application of effective negative feedback to a system - product of experts allows the negative feedback to have maximum impact on an undesired region of the state space, and learning only the negative policy \(\pi_\alpha\) at each feedback step minimises the learning time for a typical imitation learning method.

Algorithm \ref{alg:feedback} contains two key elements for the feedback process. The first is the selection function to retrieve the desired region of the trajectory $t_i$, which discards positive regions of the trajectory. This is included to eliminate, for example, the initial and final trajectory segments, where every trajectory diverges from the starting point or converges on the ending point. If this step is not taken, the success rate of the feedback system drops dramatically, as trajectories are pushed away from known regions of space, into unknown ones. 

Different selection methods, including manual selection, may be applicable for different tasks.

In this work, an effective automatic selection method is proposed, which is a simplified algorithm analogous to Ant Colony Optimisation (ACO) \cite{antColonyOpt}.

\begin{figure} [htbp!]
    \centering
      \includegraphics[width=.6\linewidth]{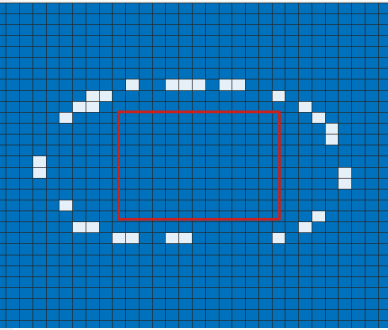}
    \caption{\emph{An illustration of the mask for an obstacle avoidance task, with zeroes, regions where $N_{threshold}$ or greater trajectories pass through, shown in white and ones, where fewer than $N_{threshold}$ trajectoreis pass through, shown in blue, and the obstacle being avoided shown in red.}}
    \label{fig:maskReal}
\end{figure}

Functionally, the state space is discretised into segments corresponding to the discretised probability distribution. Each discretised block has a certain number of trajectories, $N_t$, passing through it. If $N_t > N_{threshold}$, for a value of $N_{threshold}$ expressed here as a percentage of the total number of trajectories, then the block in question is set to a value of zero, otherwise it has a value of 1. This produces a masking distribution, $\mu$, which prevents the negative feedback influencing regions that a given percentage of the demonstrations pass through. This can be seen in Figure \ref{fig:maskReal}.

Figure \ref{fig:maskReal} illustrates the construction of a masking distribution for a simulated obstacle avoidance task, with a state space discretised into 30 blocks and an $N_{threshold}$ of 50 percent. As can be seen, for this relatively simple case with an ambiguity of $A=2$, large sections of the trajectories above and below the obstacle are overlaps between all trajectories on those paths. The masking distribution, once applied to the probability density function, will prevent negative feedback from affecting the areas in white, allowing it only to affect the regions in blue.

A higher $N_{threshold}$ indicates less confidence in the optimality of the trajectory dataset - for a dataset that is completely optimal, $N_{threshold}$ may be set such that any segment of the trajectory containing a trajectory cannot be altered by the negative feedback.

\section{Results} \label{results}
This section presents the results of the novel algorithm outlined in this paper. Subsection \ref{simulation} presents these for two simulated ambiguous tasks, a simple obstacle avoidance task and a more complex 'slalom' obstacle avoidance task; while subsection \ref{realrobot} presents the results for an ambiguous pick-and-place task implemented on a real robot. Subsection \ref{discussion} is a discussion of these results and their implications.
\subsection{Simulation} \label{simulation}
\subsubsection{Simple Ambiguous Task}
\begin{figure} [htbp!]
    \centering
      \includegraphics[width=.45\linewidth]{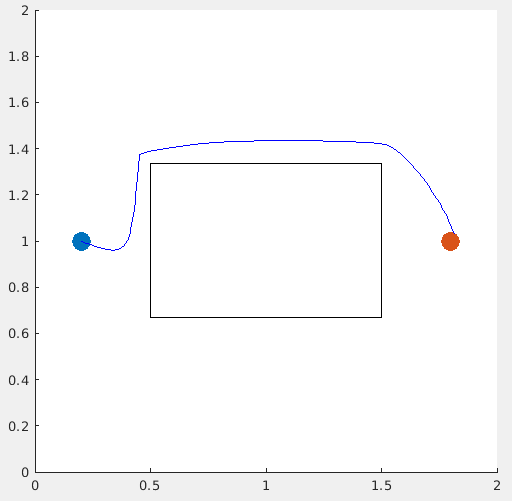}
      \includegraphics[width=.45\linewidth]{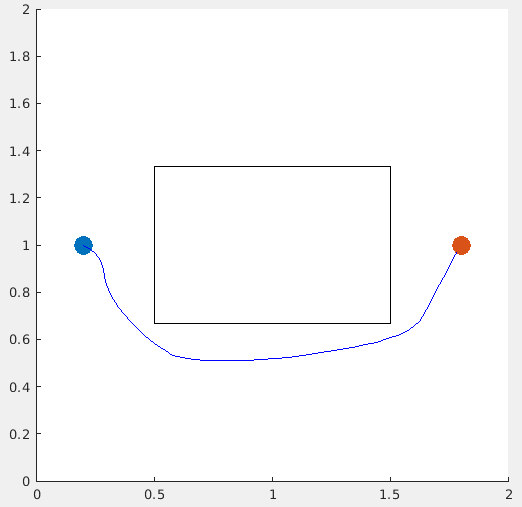}
    \caption{\emph{An illustration of the simple ambiguous task tested here, with the two success modes demonstrated, over (left) and under (right)}}
    \label{fig:taskdemo}
\end{figure}

To test this negative reinforcement system, the algorithm is implemented on a simple representative task, shown in Figure \ref{fig:taskdemo}. This Negative feedback algorithm used the methodology outlined above, compared to the same methodology using a Mixture of Experts method rather than product of experts, and to the same behavioural cloning system with a negative weight applied to failed trajectories.

Using only two success modes gives a very clear ambiguity - following the definition given in Bensch et al. \cite{Bensch10Ambiguity}, in this case $A=2$, as there are two successful behaviours, leading to two potential hypotheses: $h_a$,- going above the obstacle, and $h_b$, going below the obstacle.

In practice, the ambiguity may be higher due to inconsistent demonstrations, however, this model gives an idea of the relative simplicity of this system compared to the later experiments.

Samples of both types of demonstration (above and below the obstacle) were provided at random from a selection of ten demonstrations of each type, selected to achieve as close to an equal distribution of both types as possible. A Gaussian Mixture Regression (GMR) method sourced from Calinon \cite{Calinon07} was used to learn from the demonstrations and generate runs at the task. These runs could produce either a success, a collision failure, where the trajectory intersects with one of the obstacles; or a total failure, where the trajectory fails to reach the goal position.

\begin{figure}
\centering
\includegraphics[width = .9\linewidth]{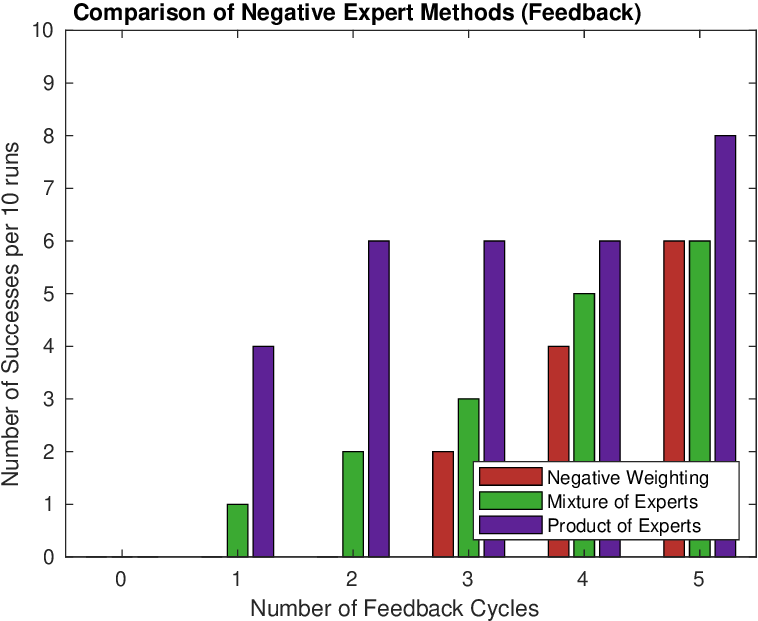}
\caption{\emph{The Negative Feedback graph for the simple ambiguous task, showing the improvement in success rate with increasing cycles of negative feedback, comparing three different methods for applying negative feedback - a negative weighting method, a mixture of experts method, and finally the product of experts method presented in this paper}}
\label{fig:simpleDemos}
\end{figure}

When there was a failure of either type, the failed output result was filtered out by a human observer, and used to learn an avoidance distribution, $\pi_{\alpha i}$, as described Section \ref{methodology}. $\pi_{\alpha i}$ was then applied through the method outlined, with a 50\% masking distribution, to the overall policy. This was tested for sample sets consisting of equal numbers of randomly selected demonstrations, first without negative learning, then with increasing numbers of cycles of negative learning; using a negative weighting method, a Mixture of Experts method, and finally the Product of Experts method presented in this paper. The Mixture of Experts method is derived from the Product of Experts method presented here, except for utilising Mixture of Experts to combine the negative distribution with the existing distribution; this is used to validate the choice of Product of Experts for the negative combination. Each was tested ten times, and the number of successes measured. These results, as shown in Figure \ref{fig:simpleDemos}, show a significant improvement in the success rate of all of these tasks with the addition of negative feedback - however, Product of experts achieves higher success rates more quickly than either of the other two methods.

\begin{figure}
\centering
\includegraphics[width = .9\linewidth]{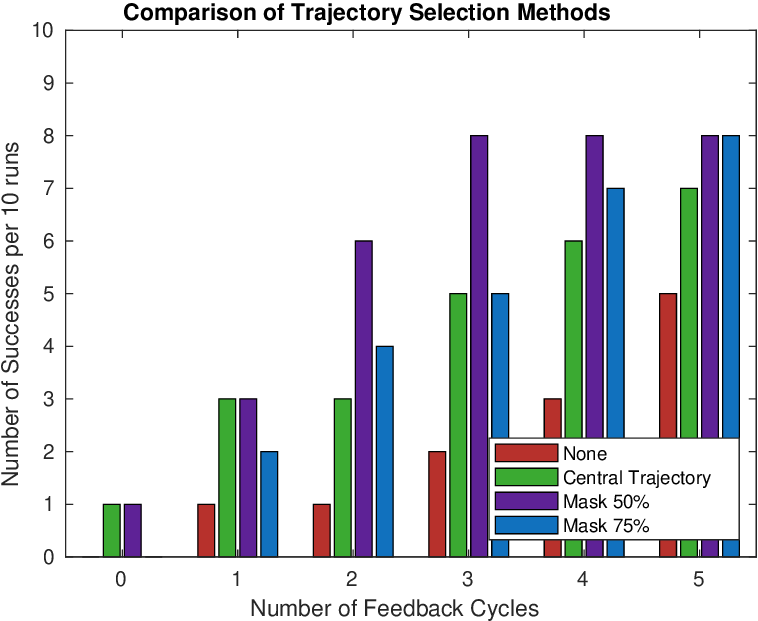}
\caption{\emph{The graph showing the impact of different methods of trajectory selection on the same five-cycle run of negative feedback described in \ref{fig:simpleDemos}. Here, 'Mask 75\%' and 'Mask 50\%' refer to the impacts of using an $N_{threshold}$ of 75\% of the trajectories and 50\% of the trajectories respectively.}}
\label{fig:trajSelectCom1}
\end{figure}

Figure \ref{fig:trajSelectCom1} shows the effects of different trajectory selection methods on the simple ambiguous task. Central trajectory refers to discarding the first and last one sixth of the trajectory, while Mask refers to the ACO-based approach this paper presents. Both methods are markedly superior to the same sequence of negative feedback with no trajectory selection - while the first round of feedback shows equivalence between central trajectory and a 50\% masking distribution, and central trajectory being marginally superior to a 75\% masking distribution, all subsequent steps of feedback show that a 50\% mask is the most effective of the trialed trajectory selection methods for this specific example.

The number of demonstrations was selected to balance the number of behaviours - if an odd number of demonstrations was used, this would bias the resultant policy towards one of the two behaviours. Essentially, for each case, the number of demonstrations per behaviour, $N_B$ is:
$N_B = \frac{N_D}{A}$
Where $N_D$ is the number of demonstrations and $A$ is the ambiguity. For both of these experiments, eight demonstrations were used for training purposes

\subsubsection{Slalom Task}

\begin{figure}
    \centering
      \includegraphics[width=.45\linewidth]{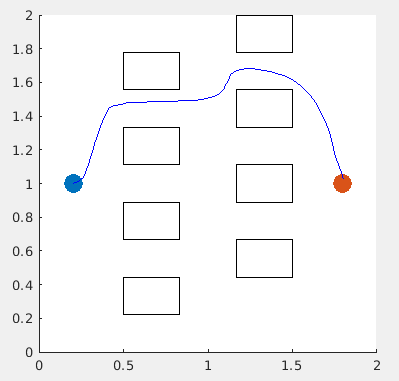}
      \includegraphics[width=.45\linewidth]{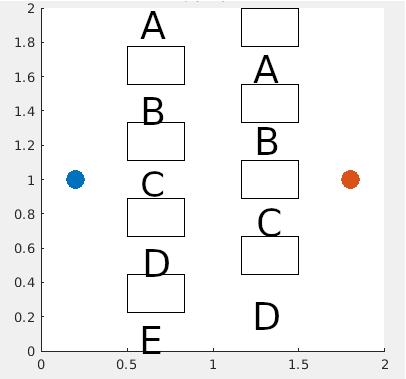}
    \caption{\emph{The 'slalom' variant of the task, which involves navigating around a more complex course, shown in the left image. The right image shows the gaps which the route can traverse, labeled from A to E and from A to D. Any combination of the first row and second row is a viable route through the slalom.}}
    \label{fig:Slalom}
\end{figure}

The algorithm was also tested on a higher ambiguity, 'slalom' variation of the task, shown in Figure \ref{fig:Slalom}. A total of twenty possible behaviours were tested, demonstrated by the labeling of gaps in the second image of Figure \ref{fig:Slalom} - paths could go through any combination of the first set of gaps with the second set of gaps. As before, demonstrations were selected at random from datasets, and evenly distributed across the possible paths, providing greater ambiguity for the system. For this experiment, no repeat demonstrations were provided, each new demonstration representing a new and different behaviour provided to the system.

In this case, demonstrations from every combination of gaps were considered: A then A, A then B, etc. With five gaps, A to E followed by four, A to D, there were twenty-five possible combinations. In this case, each demonstration was provided from a different dataset, so the number of demonstrations, $n$, represents $n$ different behaviours and therefore $n$ hypotheses.

Again, following the definition of ambiguity used previously, $A=n$ for this task, though once again, the actual ambiguity is potentially higher.

The failure modes for this task were once again collision or a failure to reach the goal.

\begin{figure}
\centering
\includegraphics[width = .8\linewidth]{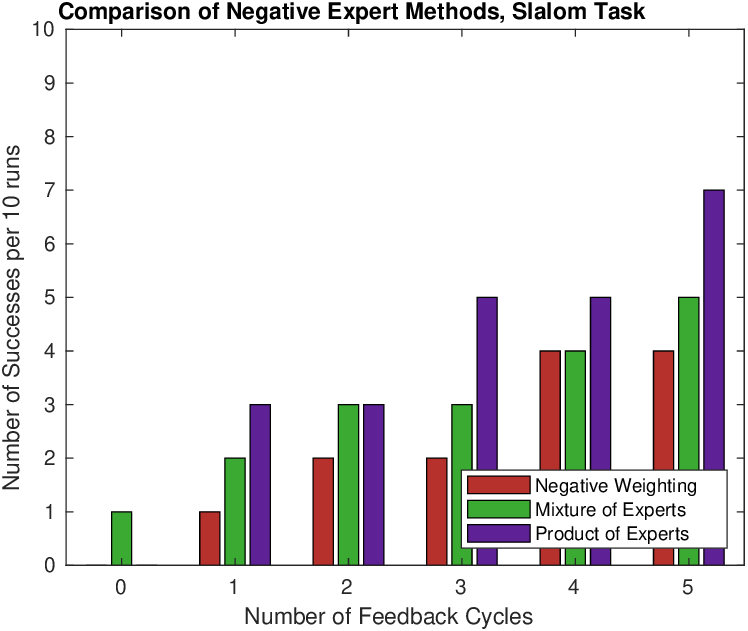}
\caption{\emph{The Negative Feedback graph for the slalom task, showing the success rate for the three different methods increasing with the number of feedback cycles.}}
    \label{fig:slalomResults}
\end{figure}

The results for the 'slalom' task, for the three chosen methods, are shown in Figure \ref{fig:slalomResults}, for five cycles of negative feedback. As can be seen from these data, for the more ambiguous task, the algorithm is less efficient, taking longer to achieve performance parity with the previous task. However, it is still capable of achieving a 50\% success rate with a product of expert method with only three cycles of negative feedback.

\begin{figure}
\centering
\includegraphics[width = .8\linewidth]{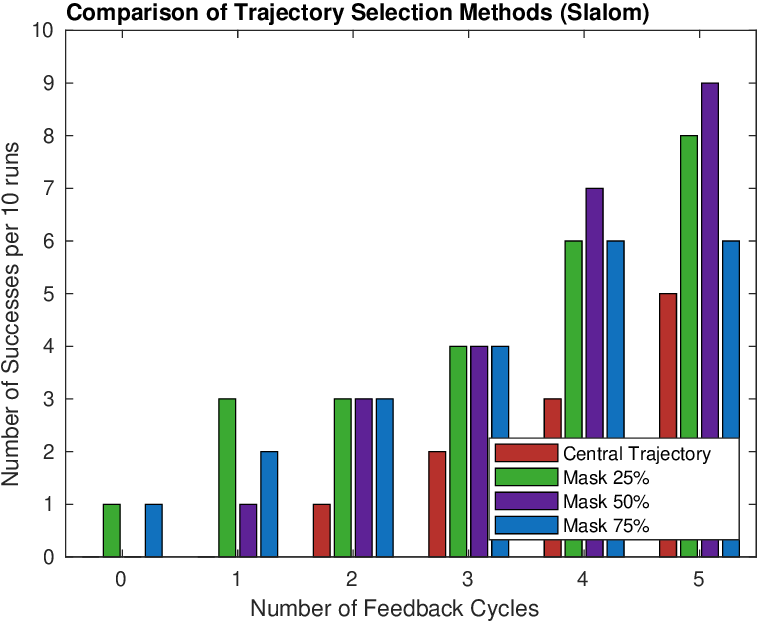}
\caption{\emph{The Negative Feedback graph for four different masking methods on the slalom task. In this instance, only central trajectory and three different masking distributions - 25\%, 50\%, and 75\% - were considered. }}
    \label{fig:trajSelectCom2}
\end{figure}

The comparison of different trajectory selection methods for the slalom task is shown in Figure \ref{fig:trajSelectCom2}, which again shows that a 50\% masking distribution is effective for this task, moreso than alternative lower or higher masking distributions. Again, the impact of the more ambiguous task is visible in the higher number of runs required for parity with the simpler task, however the improvement on baseline is still visible, even with a Central Trajectory selection scheme.

Two of the successful negative feedback runs for this task were observed to achieve their successes by traversing the central gap between the obstacles, despite having no relevant demonstrations, indicating a possibility for achieving simple tasks exclusively with suboptimal demonstrations and negative feedback.

\subsection{Real Robot} \label{realrobot}

Our system was also tested on a real robotic system, attempting a pick-and-place task in three dimensions, with an obstacle obstructing the path of the arm.

\begin{figure} [htbp!]
    \centering
    \includegraphics[width=0.8\linewidth]{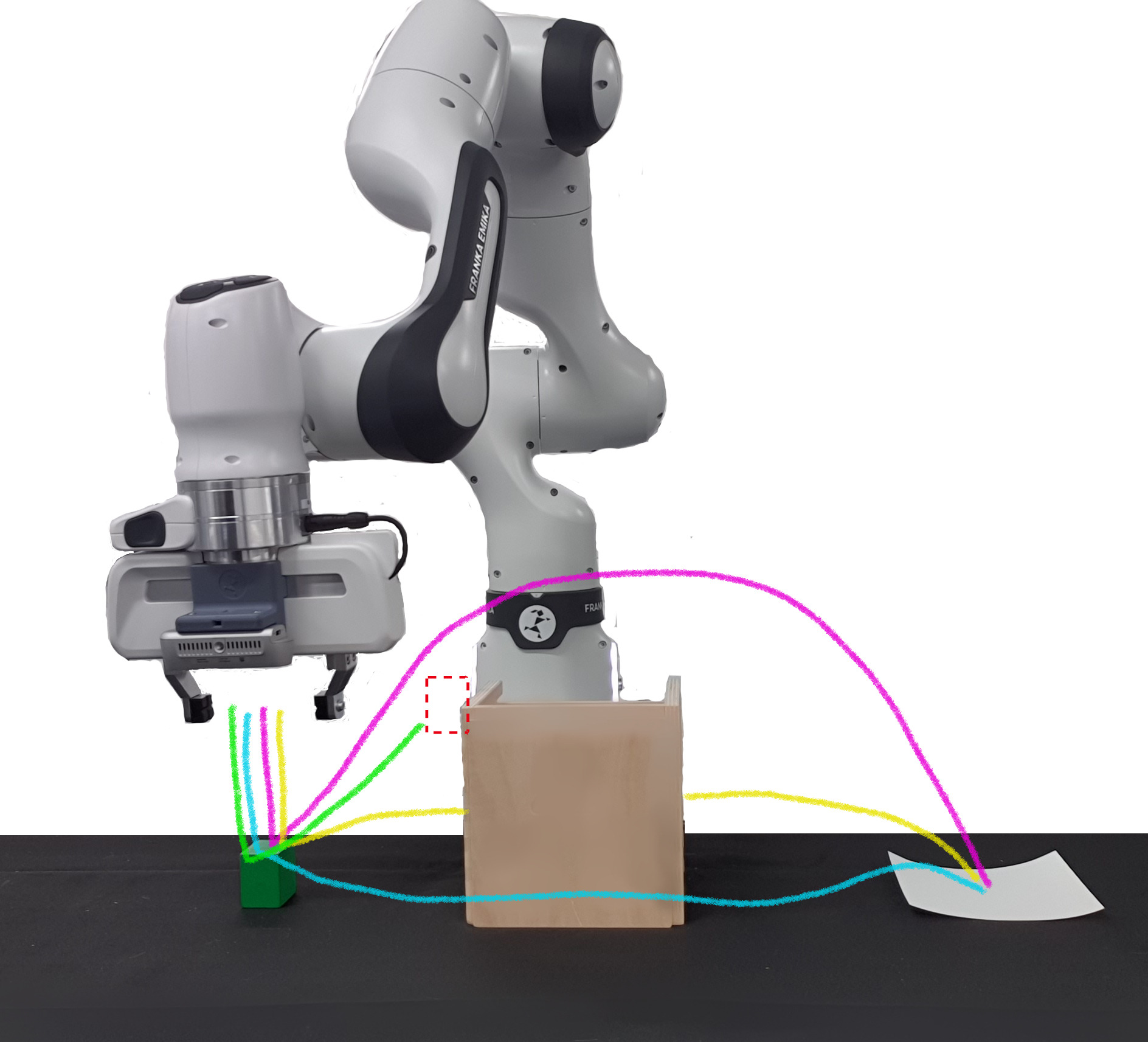}
    \caption{\emph{An illustration of the same pick-and-place with obstacle task as used in the real robot experiments, with the Franka Emika Research 3 arm visible. The highlighted lines illustrate three possible expert behaviours, and then the combination of these, which due to the ambiguity of choice between the three behaviours results in a collision.}}
    \label{fig:grabSide}
\end{figure}

The robot used was a Franka Emika Research 3 robotic arm with 7 degrees of freedom, utilising a cartesian impedance controller sourced from Franzese et al. \cite{franzese2021ilosa}. As can be seen in Figure \ref{fig:grabSide}, a small cube was used as the object for the pick-and-place task, placed on a reference point to maintain the initial position. A box was used as the obstacle, sized as approximately one third of the distance between the start and end positions. Any contact with the obstacle was considered a failure, whether it was the object that made contact, or the arm itself. In the arm's case, it was typically the sides of the gripper which collided with the box, since these extend quite far from the position of the end effector.

The ambiguity present in this task can be seen in Figure \ref{fig:grabSide}, where three possible behaviours are illustrated by three of the lines. The typical failure behaviour is also shown, a path that averages all three demonstrations and collides with the obstacle.

The methodology used was identical to the Simple Ambiguous Task experiment, only with three possible behaviours - ten demonstrations of each type were recorded, and these were selected from at random to utilise for learning.

For the ambiguity in this case, $A=3$, as there are three different behaviours possible for the system. Once again, the actual ambiguity may be higher, and especially so in this case, as it is using a three-dimensional plane where the previous tasks were only in two dimensions.

\begin{figure}
\centering
\includegraphics[width = .8\linewidth]{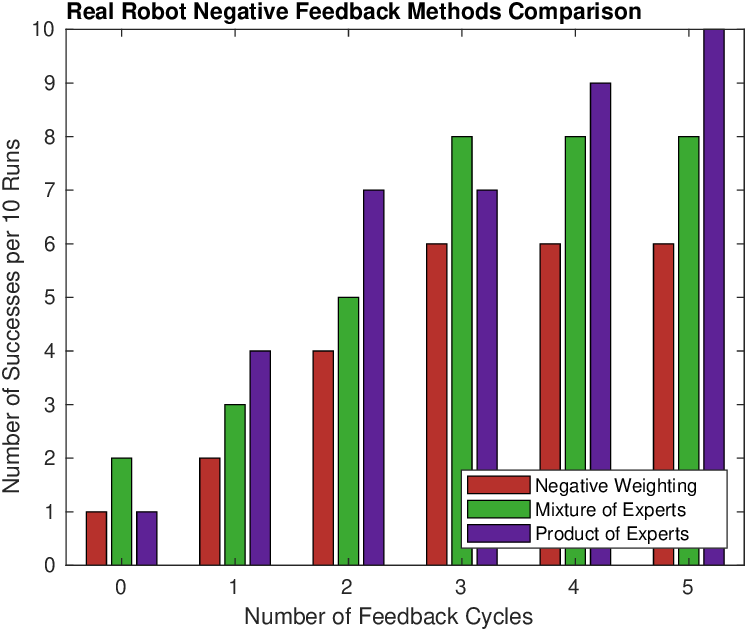}
\caption{\emph{The Negative Feedback graph for the real robot, showing the success rate per 10 runs with an increasing number of negative feedback cycles for the same three methods used in the Simulated experiments.}}
\label{fig:realNegFeedback}
\end{figure}

The results can be seen in Figure \ref{fig:realNegFeedback}, which shows the same general trend as seen in the simulation, where the negative feedback can turn very low success rates, such as 30 percent in the case of 3 demonstrations, into higher success rates, 80 percent for 3 demonstrations, within a short series of negative feedback cycles.

The success rates for this task are higher and rise faster than those for the simple ambiguous task in Figure \ref{fig:simpleDemos}, for all three methods. This is potentially due to the inclusion of the 'overhead' behaviour - with this inclusion, the average path for the system trends upwards, which is more likely to produce a successful result. While the average of left and right options will collide directly, including the over option introduces the possibility of the object passing very closely over the top of the obstacle.

\subsection{Efficiency}

One of the considerations made for this algorithm was a theoretical efficiency benefit over comparable negative learning methods specifically for negative feedback.

\begin{figure}
\centering
\includegraphics[width = .8\linewidth]{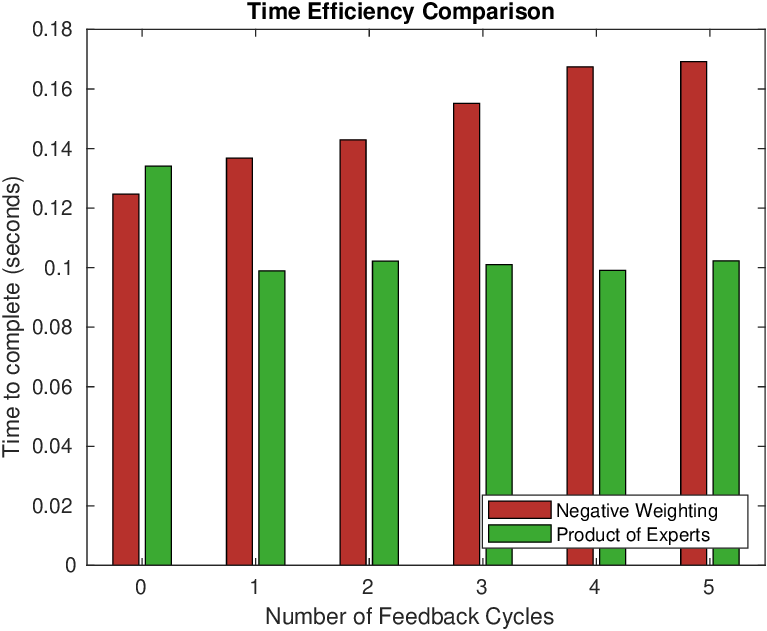}
\caption{\emph{The Negative Feedback graph for the efficiency comparison, showing the time taken to complete each step of negative feedback under the Product of Experts and Negative weighting methods, each step averaged over twenty runs.}}
    \label{fig:efficiencyResults}
\end{figure}

Figure \ref{fig:efficiencyResults} shows a comparison- of the time-to-complete between the Negative Weighting method and the novel Product of Experts approach over a run of five cycles of negative feedback, on the same hardware. While Product of Experts is marginally slower on the first step due, primarily, to the calculation of the masking distribution, once that calculation has been completed, every subsequent PoE run is significantly more efficient than comparable Negative Weighting runs.

As the number of negative feedback cycles increases, the total number of demonstrations - positive and negative - also increases. These experiments were performed with a small number of initial demonstrations to maximise failure rate, which results in an initial learning time not particularly different than the time taken for the product of experts method.

However, as the number of demonstrations increases, the time a negative weighting method takes to complete will also increase - while this will also affect the initial learning time of a product of experts based system, it will not affect subsequent iterations, as they do not incorporate data from every trajectory, only from the single failure.

In addition, a comparison of memory consumption showed that the minimal dataset used in the real robot experiments utilised 8768 Bytes of memory, while everything necessary to complete a run of the Product of Experts scheme utilised only 512 bytes. As the dataset expands, this will increase the memory consumption of the dataset but not of the Product of Experts scheme.

\subsection{Discussion} \label{discussion}

As can be seen from the data presented in Figures \ref{fig:simpleDemos}, \ref{fig:slalomResults} and \ref{fig:realNegFeedback}, negative feedback can provide dramatic improvement on success rate for imitation learning systems, and the Product of Experts method presented in this paper proves more effective for this specific application than some equivalent negative expert methods. 

The slalom task shows that this method becomes slightly less efficient with a more ambiguous task, but that it still shows significant improvement over purely positive learning, while the implementation on the real robot demonstrates the method can function both with higher dimensionality and in real-world scenarios - showing an improvement from a 10 percent success rate to a 100 percent success rate after 5 cycles of negative feedback.

Since three demonstrations refers to one demonstration of each behaviour type, this is comparable in number of demonstrations to the 2 demonstration runs of the simple ambiguous task, and achieved a more significant improvement with greater ambiguity (10 percent to 100 percent compared to 0 percent to 60 percent, with $A=3$ compared to $A=2$).

This indicates that this method could have potential for improving on ambiguous tasks with limited data, without requiring either additional data or more task-specific requirements, and may have significant potential for improving data efficiency and success rate of real robotic systems utilising imitation learning.

In addition, this method shows both time and memory based efficiency advantages over a comparable negative weighting method of negative feedback. The fact that this novel scheme does not require maintaining a stored dataset is also potentially beneficial for applications such as assistive robotics in which there may be ethical concerns around storing datasets in deployed systems.

\section{Conclusions and Future Work} \label{conclusion}

This paper introduced a novel negative feedback algorithm to improve the performance of imitation learning on small and unlabeled datasets for ambiguous tasks. The algorithm was implemented using Gaussian Means Regression in both simulation and on a real robot to perform obstacle avoidance and pick-and-place navigation tasks. These experiments covered a range of small dataset sizes and showed definitive and significant improvement in the performance of the imitation learning system on ambiguous tasks when negative feedback was applied, allowing the system to learn from its own failues.

The primary limitation of this work is the tested application - experiments were limited to trajectory-based data for navigational tasks only. The viability of this method for a broader range of tasks such as Human-Robot Interaction remain untested.

Another potential limitation is the reliance on humans in-the-loop - if the identification of failed trajectories was automated, the system might be able to automatically run the negative feedback without human intervention until a successful trajectory was generated, allowing an improvement to the time efficiency of the system, when running multiple failure trajectories on a physical robot is unnecessary, but overall the results presented in this paper are very encouraging.

Future work will include implementing a full system designed to learn from small datasets and multiple suboptimal demonstrators, as well as testing the algorithm on more complex tasks, potentially including interacting with a human.

Additionally, we intend to implement and test an automated ranking function for success and failure utilising a cost function, which would rapidly implement cycles of negative learning to improve overall performance before any runs are attempted on the physical robot.

The contribution of this work is threefold: firstly, we define a method for negative imitation learning utilising Product of Expert methods; secondly, we define an additional system of trajectory selection based on Ant Colony Optimisation in order to effectively use this method for a negative feedback algorithm; and thirdly, we validate this method for negative feedback against comparable existing methods.

This has potential to be beneficial in the context of applying imitation to, in particular, home robotics, allowing  users inexperienced with, or, in the case of assistive robotics, potentially incapable of, robotic demonstration to provide small datasets of potentially ambiguous, suboptimal data, and allowing the robot to efficiently learn from this limited data by learning from its failures.

%
%
\printbibliography
\end{document}